\documentclass[11pt]{article}

\usepackage[preprint]{acl}

 \usepackage{microtype}

\usepackage[english]{babel} 
\usepackage{newtxtext,newtxmath}
\usepackage{linguex}
\usepackage{graphicx}
\usepackage{array}
\usepackage{makecell} 
\usepackage{changepage} 
\usepackage[shortlabels]{enumitem} 
\usepackage{xfrac}
\usepackage{booktabs}
\usepackage{tabularx}
\usepackage{threeparttable}
\usepackage{fancyvrb}
\usepackage{siunitx}  

\usepackage{multirow}

\usepackage{adjustbox}

\usepackage{amssymb}
\usepackage{bm} 
\usepackage{enumitem} 

\let\emptyset\varnothing

\usepackage{color,soul} 

\title{Closing the Gap at CRAC 2026: Two-Stage Adaptation for LLM-Based Multilingual Coreference Resolution}

\author{
Antoine Bourgois
\and Olga Seminck
\and Thierry Poibeau
\\[+1.5mm]
Lattice (CNRS UMR 8094 \& ENS-PSL \& Université Sorbonne Nouvelle), Montrouge, France\\
\\[+0.5mm]
\texttt{antoine.bourgois@protonmail.com}, \\
\texttt{olga.seminck@cnrs.fr}, 
\texttt{thierry.poibeau@ens.psl.eu}\\
}

\begin{document}

\maketitle

\begin{abstract}
We present our submission to the LLM track of the 2026 Computational Models of Reference, Anaphora and Coreference (CRAC 2026) shared task. With an average CoNLL F1 score of 74.32 on the official test set, our system ranked first in the LLM track, and third overall. Our system is based on the Gemma-3-27b model, fine-tuned using a two-stage strategy with a multilingual base adapter followed by dataset-specific adapters. We represent mention spans by their headword using an XML-inspired format with local reindexing and annotate documents iteratively. These design choices proved effective across languages, document lengths, and annotation guidelines.

\end{abstract}

\section{Introduction}


Coreference resolution (CR), the task of identifying and grouping text spans (mentions) that refer to the same real-world entity, is a fundamental component of natural language understanding. It underpins downstream tasks such as information extraction \citep{yao_2019}, text summarization \citep{liu_2021}, and machine translation \citep{vu_2024}. Beyond these general NLP applications, CR has critical implications across a wide range of specialized domains, including biomedical literature \citep{lu2021}, clinical records \citep{tourille2020}, political science \citep{radford2020}, and computational humanities \citep{barre_detectives_2025}, each bringing distinct annotation conventions and linguistic challenges.

\subsection{CR Systems Evolution}

The evolution of automatic CR mirrors the broader evolution of natural language processing. Early systems in the 1970s and 1980s were primarily rule-based, relying on hand-crafted heuristics and syntactic constraints to resolve pronominal anaphora \citep{winograd1972understanding, hirst1981anaphora}.

Following the availability of large-scale annotated datasets \citep{muc_6}, the field shifted toward data-driven machine learning methods. First, statistical classifiers focused on mention-pair models \citep{soon2001machine}, then mention-ranking architectures \citep{denis2008specialized}, typically separating the task into distinct stages of mention detection and clustering. 

The introduction of deep neural models in end-to-end architectures marked another step in CR \citep{lee2017end}. The subsequent integration of Transformer-based encoders like BERT and SpanBERT \citep{joshi2019bert} led to steady improvements on benchmark datasets \citep{porada_2024}.

Solutions based on seq2seq models \citep{zhang-etal-2023-seq2seq} and generative large language models (LLMs) \citep{zhu-etal-2025-llmlink} have been introduced. These generative approaches appear promising, while also revealing significant limitations regarding the difficulty of formalizing the CR task for language models, and higher computational requirements \citep{gan-etal-2024-assessing}.

\subsection{Datasets}

CR systems have long been trained, evaluated, and optimized solely on restricted datasets such as OntoNotes, which consist primarily of news, broadcast conversation, and web data \citep{Hovy_2006}. 

As CR drew wider attention, it became evident that models trained on those generic datasets underperformed when applied to domain-specific tasks \cite{xia2021}. To address this issue, dedicated datasets have been developed, covering areas such as encyclopedic \citep{Ghaddar_2016} or biomedical data \citep{Cohen2017}, literary works \cite{Cranenburgh_2019,bamman-etal-2020-annotated,Melanie_2024} or legal documents \citep{wei2025}. Besides the development of resources for new domains, the number of languages for which resources were created also grew. 

The proliferation of specialized corpora led to a fragmented landscape, with datasets differing in annotation schemes and guidelines, file formats, and evaluation metrics ; ultimately making comparison and generalization difficult. This situation highlights the need for a unified benchmark enabling consistent evaluation across datasets.

\section{CorefUD and CRAC Shared Task}

\subsection{CorefUD Initiative}

The CorefUD initiative aims to integrate heterogeneous CR corpora within a common framework \cite{CorefUD_1_0}. It harmonizes independently developed datasets into a standardized format based on Universal Dependencies \cite{de_marneffe2021}, enabling more reliable cross-domain and cross-lingual evaluation.

In its latest version, CorefUD 1.4 comprises 33 datasets covering 19 languages and 7M tokens \citep{CorefUD_1_4}. All datasets are formatted following the CoNLL-U standard, with consistent encoding of coreference and related phenomena. The collection spans a wide range of languages, including non-European ones (e.g., Hindi, Korean) and ancient ones (e.g. Church Slavonic, Ancient Greek, Ancient Hebrew).

\subsection{CRAC Shared Task}

Built on a subset of CorefUD, the CRAC shared task provides a unified benchmark for multilingual coreference resolution. It aims to standardize evaluation and encourages the development of robust systems across datasets.

Since its first edition in 2022 \citep{CRAC_findings_2022}, the shared task has progressively expanded both in scope and difficulty. The 2024 edition introduced zero-anaphora resolution and incorporated additional datasets covering low-resource and historical languages \citep{novak2024findings}. The 2025 edition introduced a dedicated LLM-track, alongside the traditional unconstrained track, highlighting the growing interest in generative approaches \citep{novak2025findings}.

The 2026 edition features five new datasets, for a total of 27 datasets. Notably, two of these corpora (Dutch-OpenBoek \citep{nl_openboeck_2022} and French-LitBankFr \citep{Melanie_2024}) contain significantly longer documents that are on average twice the length of the longest document present in the 2025 edition of the shared task.

\subsubsection{Evaluation Metric}

Systems are evaluated using the CoNLL F1 score \citep{Pradhan_2012}, using head-based mention matching and excluding singletons. It is computed as the average of MUC, B\textsuperscript{3}, and CEAF\textsubscript{e} metrics. Final rankings are determined by the macro-average of CoNLL F1 scores over all the test sets.

\subsubsection{Previous Edition Results}

In the 2025 edition, the overall top-performing system was \textit{CorPipe}, which builds on several years of iterative improvements and refinements \citep{straka_2025_corpipe}. It relies on multilingual pretrained encoder architectures combined with careful training strategies and ensembling, and continues to set the performance standard for the shared task.

In parallel, the LLM-track participants explored a range of approaches, including fine-tuned models and few-shot or prompt-based systems. 
Within its track, \textit{GLaRef-CRAC25} \cite{seminck_2025_glaref} ranked first with a score of 62.96, yet remained substantially below CorPipe (75.84; $\Delta = -12.8$ points).

This gap highlights both the current limitations of LLM-based approaches for CR and their potential for improvement, especially given their strong performance on a wide range of other NLP tasks. In this context, our submission focuses on improving LLM-based coreference resolution.

\subsection{Contributions}

Our main contributions are:

\begin{itemize}[leftmargin=*]
    \item \textbf{Format and Task Optimization:} We introduce a minimal XML headword format, coupled with a custom cleaning function and a local reindexing strategy, which prove to be effective in practice.

    \item \textbf{Two-Stage Adaptation:} We propose a two-stage fine-tuning strategy, consisting of a robust multilingual base adapter followed by dataset-specific continual SFT to address inconsistencies in annotation guidelines across corpora.

    \item \textbf{Final Model Performance:} Our final system ranked first in the LLM-track and achieves competitive results with the best unconstrained systems on several datasets.

    \item \textbf{Open-Source Release:} We release our complete training and inference pipeline\footnote{\href{https://github.com/lattice-8094/coref-llm}{https://github.com/lattice-8094/coref-llm}}, along with the multilingual base adapter and 27 dataset-specific adapters\footnote{\href{https://huggingface.co/collections/lattice-nlp/coref-llm}{https://huggingface.co/collections/lattice-nlp/coref-llm}}, to facilitate further research in LLM-based coreference.
\end{itemize}

\section{System Development}

We develop our LLM-based coreference resolution system upon the 2025 best submission logic \citep{seminck_2025_glaref}.

The general workflow is iterative: the model annotates documents in batches of $N$ sentences at a time, feeding part of the previously annotated text into the subsequent batch. 
We use the \texttt{Gemma-3-it} family as the base model, which demonstrates strong multilingual performance and supports long contexts of up to 128K tokens \citep{gemmateam2025gemma3technicalreport}.

For all experiments and the final submission, we use parameter-efficient fine-tuning (PEFT) with 4-bit quantization via QLoRA (low-rank Adaptation) \cite{dettmers2024qlora}.

The general prompt template used for our system is illustrated below:

\medskip

\noindent\begin{minipage}{\linewidth}
\smallskip
\noindent\rule{\linewidth}{0.2pt}
\smallskip
{\small\ttfamily
\noindent\textbf{TASK:} COREFERENCE ANNOTATION\\
Annotate mentions and zero anaphora. Do not modify the input text.

\smallskip
\noindent\textbf{ALLOWED TAGS}\\
- Entities: \{OpenTag\} <EntitySpan> \{CloseTag\}\\
- Zeros: <ZeroMentionHead> \{ZeroNodeTag\}

\smallskip
\noindent\textbf{PREVIOUS CONTEXT}\\
\{250 annotated tokens\}

\smallskip
\noindent\textbf{INPUT TO ANNOTATE}\\
\{4 unannotated sentences\}

\smallskip
\noindent\textbf{ANNOTATED OUTPUT}\\
\{model output\}
}\\
\noindent\rule{\linewidth}{0.2pt}
\end{minipage}

\subsection{Scaling Law}

To accelerate iterative experimentation, we refine this base configuration using a smaller 1B-parameter model, rather than the 27B model intended for the final submission. 

Following the findings of \citet{kaplan2020} neural scaling laws, we assume that improvements observed on this proxy model will transfer to the larger one. This assumption is expected to hold for pre- and post-processing steps, annotation format choices, and dataset-specific continual supervised fine-tuning (SFT). However, it does not hold for hyperparameters choice such as learning rate, batch size, or optimal training epoch, which are inherently model-scale dependent.

Regarding the iterative annotation configuration, the \texttt{Gemma-3-1B-it} model annotates up to 4 sentences per pass with a maximum previous context of 250 words. This configuration is used both for training and inference.

\subsection{Baseline Model}

We employed the \texttt{text2text-coref} tool\footnote{\url{https://github.com/ondfa/text2text-coref}}, provided by the CRAC organizers, to convert CoNLL-formatted datasets into plaintext with in-line annotations, and to clean the model’s plaintext output and convert it back into CoNLL-U format.

The baseline model is trained for a single epoch on the concatenation of all the datasets.

On the development set this model reaches an average CoNLL F1 score of 48.22. Results differ substantially across corpora. Whereas for some datasets we observe scores above 60 points, for others the system's performance is poor (as low as 4.93 for the Latin dataset).

From this baseline, we refine our approach and evaluate the impact of the proposed modifications.

\subsection{Annotation Format}

\begin{table*}[htbp]
\centering
\setlength{\tabcolsep}{3pt}
\renewcommand{\arraystretch}{1.15}
\renewcommand{\tabularxcolumn}[1]{m{#1}}

\begin{threeparttable}

\begin{tabularx}{\textwidth}{m{2.5cm} X >{\centering\arraybackslash}m{1.5cm}}
\toprule
\multicolumn{1}{l}{\textbf{FORMAT}} & \multicolumn{1}{c}{\textbf{TEXT}} & \multicolumn{1}{c}{\textbf{TOK}} \\
\midrule
Input & When Lison visits her sister , $\emptyset$\tnote{a} brings flowers. & 10 \\[5pt]
\midrule
CRAC & When Lison\textbf{|[e1]} visits her\textbf{|[e1],[e2} sister\textbf{|e2]} , brings \textbf{\#\#|[e1]} flowers. & 29 \\[7pt]
Explicit XML & When \textbf{<ent id=COREF\_1>} Lison \textbf{</ent>} visits \textbf{<ent id=COREF\_2>} \textbf{<ent id=COREF\_1>} her \textbf{</ent>} sister \textbf{</ent>} , brings \textbf{<zero\_ent id=COREF\_1>} flowers. & 56 \\[7pt]
Minimal XML & When \textbf{<ent1>} Lison \textbf{</ent>} visits \textbf{<ent2>} \textbf{<ent1>} her \textbf{</ent>} sister \textbf{</ent>} , brings \textbf{<zero1>} flowers. & 35 \\[5pt]
\midrule
Headword XML & When Lison \textbf{<ent1>} visits her \textbf{<ent1>} sister \textbf{<ent2>} , brings \textbf{<zero1>} flowers. & 26 \\
\bottomrule
\end{tabularx}

\begin{tablenotes}
\footnotesize
\item[a] $\emptyset$ null subject of “brings”.
\end{tablenotes}
\vspace{-0.5em}
\caption{A made-up example in English featuring a zero-mention to illustrate various coreference annotation formats, from raw text to explicit XML and mention-head marking. Token counts (TOK) indicate how many subword tokens each format produces. Subwords are tokenized using the \texttt{Gemma-3} tokenizer.}
\label{tab:plaintext_formats}
\end{threeparttable}
\vspace{-1em}
\end{table*}

The first direction for improving the LLM-based CR concerns the plaintext format provided to the LLM, and more precisely the inline annotation scheme.

The format provided in the CRAC shared task uses a compact plaintext encoding, where mention boundaries and entity identifiers are embedded inline using bracket markers. While this format is token-efficient as it uses one tag for single-token mentions; as suggested by \citet{seminck_2025_glaref}, it might not be the most interpretable for LLMs. 

We explore alternative tagging schemes inspired by markup languages, which are likely well represented in the model’s pretraining data. These clearly delimit mention boundaries with readable, nested tags, explicitly marking start and end of each span (\texttt{<entity\_start> ... </entity\_end>}).

Furthermore, while the CRAC format repeats the coreference chain identifier in the closing tag (for multi-token mentions), and since mention span boundaries cannot cross, we can take advantage of the “last open, first closed” principle, which allows us to represent nested mentions without repeating the coreference indices, as the ID can be implicitly recovered from the most recently opened tag. For the rare discontinuous mentions, we retain only the fragment containing the mention head.

\autoref{tab:plaintext_formats} illustrates the two XML-based formats we experimented with:

\begin{enumerate}
    \item \textbf{Explicit XML} annotations, where each mention span is wrapped in fully specified tags (e.g., \texttt{<ent id=COREF\_1>}). This format makes entity boundaries and identities unambiguous and explicitly structured. For closing boundary, a generic \texttt{</ent>} tag is used. Zero mentions are marked by inserting a dedicated tag (\texttt{<zero\_ent id=COREF\_1>}) after the syntactic head.
    \item \textbf{Minimal XML}, where entity tags are shortened (e.g., \texttt{<ent1>}, \texttt{</ent>}, \texttt{<zero1>}) to reduce verbosity. This preserves most of the structural clarity of the explicit XML format while lowering the tokenization overhead.
\end{enumerate}

The impact of these design choices is reflected both in tokenization cost and downstream performance. As shown in \autoref{tab:plaintext_formats}, more explicit formats substantially increase the number of subword tokens: 56 tokens for Explicit XML and 35 for Minimal XML versus 29 for CRAC plaintext. Experimental results with \texttt{Gemma-3-1B-it} (\autoref{tab:plaintext_formats_results}) indicate that annotation format has a measurable effect on model performance. XML-based approaches show contrasting behaviors. The Explicit XML format leads to a noticeable drop in performance (45.33), suggesting that increased verbosity and longer tag structures may hinder the model’s ability to effectively process the input. In contrast, the Minimal XML variant achieves the best overall performance (51.03), outperforming the CRAC format by +2.81 points.

This suggests that improving the structural clarity of annotations can benefit LLM-based CR, but only when balanced with token efficiency. Overall, this experiment highlights the sensitivity of LLMs to input representation and emphasizes that annotation design is critical.

\begin{table}
\centering
\begin{tabular}{l S[table-format=2.2]@{\hspace{0em}}l}
\toprule
\textbf{Tag} & {\textbf{Avg CoNLL F1}} & {} \\
\midrule
CRAC (baseline) & 48.22 & {} \\
Explicit XML & 45.33 & (-2.89) \\
Minimal XML & \textbf{51.03} & (+2.81) \\
\bottomrule
\end{tabular}
\vspace{-0.5em}
\caption{Comparison of the performance of \texttt{Gemma-3-1B-it} using the different inline coreference tagging formats. Deltas indicate change relative to CRAC baseline. Details on performance per dataset can be found in \autoref{sec:development_set_formats_rsults}.}
\label{tab:plaintext_formats_results}
\vspace{-1em}
\end{table}

Adopting these XML-based annotation formats required us to rewrite the scripts for converting between the CoNLL-U and the plaintext formats. We also had to adapt the cleaning procedures applied to model outputs, to ensure that annotations are correctly recovered and aligned with the original input. These steps are necessary to reliably project predictions back into CoNLL-U format and enable proper evaluation.

\subsection{Cleaning Function}

The cleaning function provided in CRAC relies on a word-level edit distance to align generated outputs with the original input, and operates at the document level.

\subsubsection{Custom Cleaning Function}

We propose a new cleaning procedure designed to better handle the variability of LLM-generated outputs. The process consists of three main steps: (i) linking each annotation tag with an output token, (ii) aligning output tokens to the input sequence, and (iii) projecting annotation tags onto the corresponding input tokens.

To align LLM-generated text with input tokens, we implement a hierarchical anchoring strategy that mitigates lexical drift and hallucinations:

\paragraph{Recursive Anchoring}
We first establish a monotonic alignment by identifying anchor tokens that are unique in both the input and the predicted output. The alignment is then refined recursively by introducing local anchors, tokens that are unique within the unmatched regions between previously aligned anchors, allowing for progressively finer alignment.

\paragraph{Island Expansion and Fuzzy Matching}
Starting from these anchors, matches are expanded in both directions to recover contiguous spans. Remaining unmatched regions are resolved using fuzzy matching to handle minor discrepancies in surface forms.

\paragraph{Span Projection}
Annotation tags are projected onto the aligned input tokens based on the computed mapping. A stack-based parsing ensures that overlapping mentions are converted into properly nested spans, preventing invalid crossing structures.

\medskip

Our cleaning procedure operates on sequences of arbitrary length, ranging from individual sentences to full documents. This flexibility is particularly useful in our setup, where documents are processed iteratively in batches of four sentences. Cleaning is therefore applied at the batch level rather than on complete documents.

Moreover, this design enables on-the-fly cleaning during inference, avoiding the need for a dedicated post-processing step after full document generation.

Prior work in the 2025 shared task reports several issues in LLM-generated outputs that hinder alignment and evaluation, including sequence looping \citep{seminck_2025_glaref}, repetition of empty nodes \citep{hejman2025}, and other deviations from the expected annotation format \citep{phuc2025}. Our custom cleaning function solves these issues by ensuring that the returned text is identical to the input (except for annotations that have been added).

\subsubsection{On-the-fly Cleaning}

In theory, on-the-fly cleaning helps prevent corrupted intermediate outputs from being propagated across batches. We evaluate the impact of performing cleaning during inference. 

On average, on-the-fly cleaning yields a marginal improvement (+ 0.01 points) on the development set. 
The overall effect remains limited. Qualitative analysis indicates that severely corrupted annotations are relatively rare, reducing the potential impact of this cleaning strategy. Nonetheless, it does not hurt the performance so we keep on-the-fly cleaning for all subsequent experiments.

\medskip

Another problem encountered by last year participants relates to the reuse of incorrect coreference IDs. For example, if the entity ID \texttt{'2'} has already been used for a coreference chain earlier in the document, but no mention of the chain appear in the previous context, the model is susceptible to reuse this ID for a newly introduced entity as it does not have a cache of used IDs. Several solutions have been proposed to mitigate this problem, including expanding the context to cover the entire annotated document, explicitly providing the model with the set of available identifiers for new entities, or maintaining an external cache of existing coreference chains and incorporating it into the model input. In this work, we propose a local reindexing strategy.

\subsection{Reindexing Coreference IDs}

At each inference step, the coreference IDs that are visible in the previous context are remapped, so the first entity appearing in context is always assigned ID \texttt{0}, the second ID \texttt{1}, and so on up to \texttt{N}. The model is then expected to annotate new entities introduced in the current chunk using IDs in the range N+1 to N+1+E, where E is the number of newly introduced chains. This keeps the set of valid IDs small and contiguous regardless of how many entities have appeared globally throughout the document. A bidirectional mapping is maintained between these local indices and the true global IDs, so that after generation the predicted chunk is reprojected back into the global coreference space before being appended to the document-level annotation. This mapping implies negligible overhead at inference time.

Beyond reducing ID sparsity, the approach also yields a cleaner training signal: since IDs outside the visible context window are reindexed as if they were new entities, the model is never expected to recall a chain that is not in the visible context.

Applying the reindexing strategy on top of our strongest configuration (minimal XML tags with on-the-fly cleaning) yields a consistent improvement in performance, increasing the average CoNLL F1 score from 51.04 to 51.44 (+0.40) (see \autoref{sec:development_set_formats_rsults} for detailed results). While the gain is moderate, it is stable and comes at negligible computational cost. 

Reducing the sparsity and range of coreference IDs simplifies the model’s prediction space: constraining valid IDs to a small, contiguous range limits erroneous ID reuse and improves annotation consistency. These results show that even lightweight structural constraints can lead to measurable improvements in LLM-based coreference resolution.


\subsection{Headword Mentions}
Another direction for improving LLM-based coreference resolution relates to the representation of mention spans. Building on previous work \citep{dobrovolskii2021, prazak2024}, participants in last year’s shared task \citep{hejman2025} proposed representing mentions using only their syntactic head, reporting substantial gains over full-span representations. This approach is particularly well suited to the shared task setting, as the official evaluation metric relies solely on head matching.

Following this insight, we design a custom plaintext format derived from minimal XML, where a single tag is inserted immediately after the headword of each mention (e.g., \texttt{<ent1>}) or zero mention (e.g., \texttt{<zero1>}), as illustrated in \autoref{tab:plaintext_formats}.

This representation shortens input sequences while preserving the coreference signal, simplifying the learning problem in two ways. First, by reducing the number of inserted tags, it limits noise related to exact span boundaries during training. Second, it largely removes the complexity of nested mentions, since the few remaining cases where multiple mentions share the same head are handled by consecutive tags attached to that headword.

Overall, the headword representation achieves the best performance, reaching an average CoNLL F1 score of 54.40, compared to 51.44 for the previous strongest configuration. This +2.96 improvement is the largest gain observed across all tested configurations (see \autoref{sec:development_set_formats_rsults} for detailed results).

This result highlights the effectiveness of simplifying mention representations: by reducing sequence length while preserving essential coreference cues, the model benefits from a simpler and more focused prediction space, leading to more consistent and accurate annotations.

\subsection{Inter-Dataset Variability and Errors}

While these results are encouraging, with an average CoNLL F1 score of 54.40 compared to 48.22 for our baseline, performance remains highly uneven across datasets. Scores range from 0.00 on the Latin dataset to 71.75 on Deutsch, revealing substantial variability depending on language and data conditions.

A first clear trend is that datasets combining low training resource and ancient languages (e.g., Latin, Ancient Hebrew, Old Church Slavonic) consistently underperform. In contrast, well-resourced modern languages such as English, French, and Spanish achieve much stronger and more stable performance, typically around 60 CoNLL F1.

Beyond data size, annotation heterogeneity also plays a critical role. Some languages include multiple datasets with markedly different annotation guidelines. This is particularly evident in French, where \textit{fr\_democrat} and \textit{fr\_litbank} contain similar types of texts but differ substantially in annotation density. Measured in mentions per 100 tokens, their distributions range from 13.6 (litbank) to 27.9 (democrat). In fact, \textit{litbank} only annotates a restricted subset of entity types, whereas \textit{democrat} follows a more exhaustive scheme covering all referring expressions.

Although prior work \citep{hejman2025} hypothesized that models can implicitly adapt to dataset-specific conventions, our results suggest this ability is limited when datasets are closely related but follow conflicting guidelines. A similar pattern is observed in other languages, where annotation density also varies widely. Overall, mention density ranges from 8 to 38 mentions per 100 tokens across datasets, with an average of 22, highlighting the extent of cross-dataset inconsistency.

This inconsistency is reflected in the model’s predictions. On average, the model underpredicts mentions (18 predicted vs. 21 in gold annotations per 100 tokens), and tends to regress toward a global average rather than matching dataset-specific distributions (see \autoref{sec:mention_density} for details). For instance, in French, the model underpredicts on \textit{fr\_democrat} ($-28\%$) while overpredicting on \textit{fr\_litbank} (+64\%), virtually averaging the two annotation schemes.

These observations suggest that, despite strong overall performance, the model struggles to capture dataset-specific annotation conventions. Instead, it learns a smoothed, global notion of coreference structure, which leads to systematic errors when annotation guidelines diverge.


One way to mitigate this issue is to explicitly indicate which dataset is being processed as part of the input to the model. Another approach is to continue training the low-rank adapter separately for each dataset. We explore this in the next section.

\subsection{Dataset-Specific Adapters}

We build on the multilingual adapter obtained after one epoch of training on the union of all datasets, and further specialize it by continuing LoRA fine-tuning independently for each dataset for an additional N epochs.

We experimented with several values of N, and found that performance typically improves for a small number of epochs (1-5) before overfitting, the optimal value is dataset-dependent, reflecting differences in size and annotation complexity. This procedure consistently improves performance across datasets. 

\autoref{sec:DatasetAdapter} reports the performance obtained by continuing adapter fine-tuning for up to five dataset-specific epochs.

Performance improvements are particularly pronounced for smaller datasets. For instance, \textit{hbo\_ptnk} improves by +34.95 points and \textit{la\_coreflat} by +34.02, highlighting the benefit of specialization when limited training data is available. Mid-sized datasets such as \textit{lt\_lcc} and \textit{hu\_szeged} also show substantial gains (+16.57 and +11.54 respectively), while larger datasets tend to exhibit more moderate, but still consistent improvements.

The optimal number of dataset-specific epochs before overfitting varies across datasets. Some datasets peak early (e.g., one epoch for \textit{hi\_hdtb}, \textit{fr\_ancor}), while others benefit from longer fine-tuning (up to five epochs for \textit{la\_coreflat}). This variability suggests that early stopping should ideally be tuned per dataset. Overall, this strategy yields improvements across all datasets, with an average gain of +8.15 points.

Finally, we compare this approach to a uniform strategy where a single adapter is fine-tuned on all datasets for five epochs, ensuring a fair comparison in terms of compute and total exposure to training data. While this setting already improves over the multilingual baseline (54.40 -> 60.23), it remains slightly below the best dataset-specific configuration (61.37), corresponding to roughly one additional epoch of effective data exposure.

In practice, this results in a collection of 27 dataset-specific adapters, each derived from the same multilingual initialization. This introduces additional storage requirements, but adapters are lightweight, only requiring ~2 GB, consisting solely of the low-rank weights and configuration.

This strategy is also promising for adapting to new datasets or specific annotation guidelines. For instance, some downstream applications in Computational Literary Studies restrict coreference annotations to specific mention types, such as characters in literary coreference \citep{bourgois2026}. Targeted adapter fine-tuning can incorporate these constraints efficiently, without training on all datasets, since the coreference knowledge is already captured in the multilingual base adapter.


\section{Final Model}
Building on insights from experiments with \texttt{gemma-3-1b-it}, we train our final system by replacing the 1B base model with the more powerful \texttt{gemma-3-27b-it}.

All training and inference are performed on two 48GB NVIDIA RTX 6000 GPUs. We adopt the QLoRA fine-tuning setup described in the previous sections. The training procedure consists of one initial epoch on the concatenation of all datasets to learn a shared multilingual coreference representation, followed by up to three dataset-specific fine-tuning epochs. The best checkpoint is selected separately for each dataset. We retain the headword-based annotation scheme with XML-style tags, along with on-the-fly output cleaning and local reindexing during both training and inference.

\begin{figure*}[h] 

\section{Distribution of Distance to Last Coreferential Antecedent}
\label{sec:antecedent_distance}

\centering
\includegraphics[width=\textwidth]{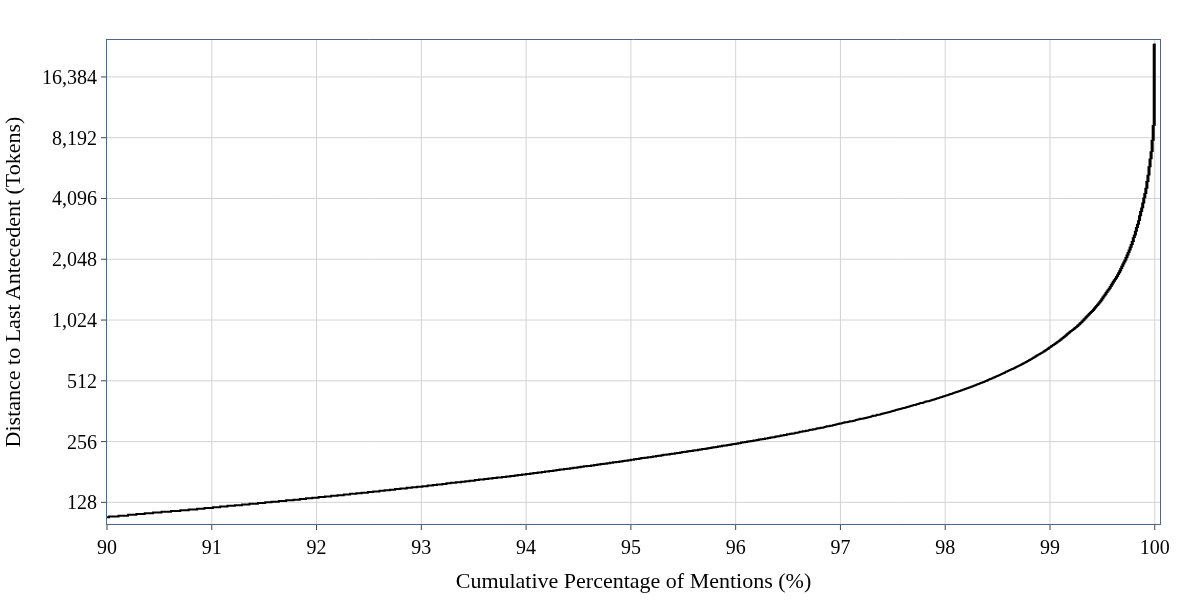}
\caption{Cumulative percentage of mentions vs distance to last antecedent. Training and development sets. 250 tokens of previous context allow covering 96\% of last coreferential antecedent. 3,072 tokens cover 99.84\%}
\label{fig:distance_to_last_antecedent}
\end{figure*}

Compared to the 1B setup, we adjust the context configuration to better leverage the larger model capacity. During training, we process batches of six sentences and use a context window of 1,024 tokens for the preceding context. At inference time, we extend this context to 3,072 tokens, which ensures that, for almost all mentions (99.84 \%), the most recent coreferential antecedent is available in the input context. This is supported by the distribution of antecedent distances reported in the \autoref{sec:antecedent_distance}.

The full fine-tuning takes approximately 160 hours. At inference time, processing the development set requires around 30 hours. See \autoref{sec:hyperparameters} for more details on hyperparameters.

\subsection{Result on Development Set}

Our system achieves strong and consistent performance across a wide range of datasets, with an average score of 75.64. This places it first in the LLM track and third overall, behind the two strongest systems from the unconstrained track.

Compared to last year's edition \citep{novak2025findings}, where the CorPipe ensemble dominated 18/22 datasets and the best LLM-based system lagged far behind ($-12.88$ points on average), our results show a dramatic reduction of the performance gap (\autoref{tab:devset_final_results}). The difference between the best unconstrained system and our model is only -1.94 points, highlighting the impact of scaling and improved pre-, post-process and fine-tuning strategies.

\begin{table}[!t]
\small
\renewcommand{\arraystretch}{1.2}
\resizebox{\columnwidth}{!}{
\begin{tabular}{lcclc}

\toprule
Track &\multicolumn{2}{c}{Unconstrained} &\multicolumn{2}{c}{LLM} \\
\cmidrule(lr){2-3}\cmidrule(lr){4-5}
\multirow{2}{*}{System} &\multicolumn{2}{c}{CorPipe} &\multirow{2}{*}{\makecell{\textbf{Ours}\\\textbf{Best Adap.}}} &\multirow{2}{*}{Hejmanj} \\
\cmidrule(lr){2-3}
& Ensemble & Single & & \\

\midrule
ca\_ancora &\textbf{85.50} &84.57 &\phantom{000}78.72 &\ul{83.44} \\
cs\_pcedt &\textbf{79.59} &78.89 &\phantom{000}74.66 &\ul{75.59} \\
cs\_pdt &\textbf{81.98} &81.58 &\phantom{000}77.15 &\ul{78.88} \\
cs\_pdtsc &\textbf{76.64} &76.11 &\phantom{000}70.70 &\ul{73.45} \\
cu\_proiel &\textbf{67.89} &66.73 &\phantom{000}\ul{60.69} &57.26 \\
de\_potsdam &\ul{81.05} &77.92 &\phantom{000}81.69\textsuperscript{**} &\textbf{82.09} \\
en\_fantasy &\ul{82.77} &81.58 &\phantom{000}\textbf{84.54} &82.77 \\
en\_gum &\textbf{79.25} &78.61 &\phantom{000}78.61\textsuperscript{**} &\ul{79.18} \\
en\_litbank &83.25 &\ul{83.76} &\phantom{000}\textbf{85.21}\textsuperscript{*} &84.11 \\
es\_ancora &\textbf{85.18} &84.56 &\phantom{000}78.70 &\ul{82.57} \\
fr\_ancor &\textbf{80.88} &79.02 &\phantom{000}\ul{80.68} &77.74 \\
fr\_democrat &\textbf{76.12} &75.15 &\phantom{000}\ul{76.09} &70.34 \\
fr\_litbankfr &\ul{79.65} &79.31 &\phantom{000}\textbf{80.33}\textsuperscript{**} &62.86 \\
grc\_proiel &\textbf{80.64} &80.28 &\phantom{000}\ul{75.79} &74.96 \\
hbo\_ptnk &\ul{73.20} &71.68 &\phantom{000}\textbf{76.63}\textsuperscript{*} &70.95 \\
hi\_hdtb &\textbf{81.74} &81.51 &\phantom{000}80.36 &\ul{81.23} \\
hu\_korkor &67.83 &\textbf{68.45} &\phantom{000}66.66\textsuperscript{**} &\ul{68.16} \\
hu\_szeged &\textbf{73.02} &72.28 &\phantom{000}67.74\textsuperscript{*} &\ul{71.88} \\
ko\_ecmt &\textbf{70.54} &69.93 &\phantom{000}65.09\textsuperscript{**} &63.77 \\
la\_coreflat &60.22 &58.54 &\phantom{000}\ul{60.84} &58.06 \\
lt\_lcc &\ul{80.45} &79.30 &\phantom{000}78.13 &\textbf{81.11} \\
nl\_openboek &74.51 &\ul{75.05} &\phantom{000}\textbf{77.25} &67.50 \\
no\_bokmaal &81.10 &\ul{81.16} &\phantom{000}80.83 &\textbf{81.72} \\
no\_nynorsk &\ul{80.52} &79.20 &\phantom{000}83.39 &\textbf{84.40} \\
pl\_pcc &\textbf{81.18} &80.49 &\phantom{000}\ul{78.49}\textsuperscript{**} &78.31 \\
ru\_rucor &\textbf{81.60} &80.80 &\phantom{000}77.30\textsuperscript{**} &\ul{78.43} \\
tr\_itcc &\textbf{68.39} &66.43 &\phantom{000}\ul{67.56} &59.39 \\
\midrule
Average &\textbf{77.58} &76.77 &\phantom{000}\ul{75.64} &74.16 \\
Std. Dev. &6.19 &6.33 &\phantom{0000}6.93 &8.48 \\
\bottomrule
\end{tabular}
}
\caption{Results on the development set with \texttt{gemma-3-27b-it}. For our submission, we report the best dataset-specific checkpoint for each corpus. ({*}) and ({**}) indicate that the best performance was obtained at the first and second training epochs, respectively; otherwise, results correspond to the third epoch. Only the top two systems from each track are shown (total 10 systems). Overall best scores in bold, best results within each track underlined (when not already bold). \textit{la\_coreflat} is the only dataset for which a fifth system (\textit{thmorton}) achieves the best score.}
\label{tab:devset_final_results}
\vspace{-1em}
\end{table}

At the dataset level, our system achieves the best score on 6 of the 27 datasets and ranks second on 4 others, demonstrating broad competitiveness. Gains are notable on French and lower-resource or historical datasets, suggesting strong generalization across diverse conditions.

Importantly, our approach also performs well on long-document benchmarks, which are known to be particularly challenging for mention-pair models due to long-distance dependencies \citep{bourgois2025} ; a problem that was also pointed out by last year's participants \citep{phuc2025}. Among the longest datasets, we rank first on nl\_openboek, fr\_litbankfr, en\_litbank and en\_fantasy, indicating that our combination of iterative decoding, local reindexing, and extended inference context (3,072 tokens) effectively captures long-range coreference links.

That said, performance remains somewhat uneven: we still trail the CorPipe ensemble or the second best LLM system on many datasets. This is reflected in a higher standard deviation (6.93) compared to CorPipe (6.19), indicating greater variability across corpora.

\subsection{Result on Test Set}

Due to time constraints during the evaluation phase, we were only able to train dataset-specific adapters for two epochs for the official test set submission, which is likely suboptimal. Based on prior results, we estimate that an additional third epoch of fine-tuning could yield further improvements on the test set performance.

Despite this limitation, the overall trend remains consistent with the development set. Our system achieves an average score of 74.32, ranking first in the LLM-track and third overall. We rank first on 5 datasets and second on 3. See \autoref{sec:testset_final_results} for detailed results on the test set. For a comprehensive analysis and comparison across all participating systems, we refer the reader to the findings of the shared task \citep{novak2026findings}.

These results show that LLM-based approaches closely match traditional pipelines, perform well on challenging datasets, and have largely closed the performance gap.

\section*{Conclusion}

We presented our Gemma-3-based submission to the CRAC 2026 shared task, achieving an average CoNLL F1 score of 74.32, ranking first in the LLM-track and third overall. Our iterative annotation strategy, minimal XML headword formatting, and local reindexing proved effective across diverse languages and document lengths. Additionally, dataset-specific adapters help mitigate guideline inconsistencies across corpora. Our results demonstrate that LLM-based systems can compete with traditional specialized pipelines, while identifying clear directions for further progress.

\section*{Limitations and Perspectives}
Despite these encouraging findings, several limitations remain, suggesting avenues for future work.

\subsubsection*{Context and Batch Configuration}
The length of input context and number of sentences per batch were chosen heuristically. Optimal settings may vary across datasets, especially given differences in document length and structure. Future work could explore dataset-specific context sizes to maximize performance.

\subsubsection*{Head-Only Span Representation}
Our minimal XML format annotates only the head token of each mention. While effective for the shared task, this approach limits applicability to real-world scenarios where full-span resolution may be required.

\subsubsection*{Task-Specific Training Objective}
Fine-tuning relied on standard cross-entropy loss, which treats all token outputs equally. Developing a task-specific loss that explicitly penalizes coreference errors could further improve model accuracy. As the CoNLL F1 metric is fully computable, reinforcement learning with a verifiable reward (RLVR) offers a promising avenue to directly optimize for coreference performance.

\subsubsection*{Computational Requirements}
Although QLoRA and 4-bit quantization improve efficiency, fine-tuning and inference with large models still demand substantial computational resources. Exploring smaller, specialized models or more efficient architectures could make the approach more accessible.

\subsubsection*{Future Model Exploration}
In this work, we only experimented with Gemma-3, which previously showed strong performance on coreference tasks. Future studies could explore alternative LLMs, such as Qwen3.5, Llama3, or Gemma-4, which may offer further gains.

\newpage

\section*{Funding}
This research was funded in part by PRAIRIE-PSAI (Paris Artificial Intelligence Research Institute – Paris School of Artificial Intelligence) \href{https://anr.fr/ProjetIA-23-IACL-0008}{ANR-23-IACL-0008}.\\ This work has received support under the Major Research Program "CultureLab" launched by PSL Research University and implemented by ANR with the references ANR-10-IDEX-0001.

\newpage

\bibliography{custom}

\appendix

\newpage

\begin{table*}[t!] 
\centering

\noindent\section{MiniDev set Annotation Format Experiments Results}
\label{sec:development_set_formats_rsults}

\setlength{\tabcolsep}{12pt}
\begin{tabular}{lcccccc}
\toprule
Tag & CRAC & \textbf{Explicit XML} & \textbf{XML} & XML & XML & \textbf{Headword} \\
Reindex & False & False & False & False & \textbf{True} & True \\
InferenceClean & False & False & False & \textbf{True} & True & True \\
\midrule
ca\_ancora & 55.20 & 59.23 & 61.94 & 62.47 & 61.72 & 64.68 \\
cs\_pcedt & 48.59 & 52.69 & 56.61 & 57.05 & 58.05 & 61.02 \\
cs\_pdt & 51.76 & 51.71 & 53.08 & 53.72 & 54.31 & 57.55 \\
cs\_pdtsc & 42.74 & 52.84 & 56.49 & 56.91 & 57.47 & 60.26 \\
cu\_proiel & 18.35 & 16.61 & 26.19 & 26.64 & 27.00 & 31.62 \\
de\_potsdamcc & 66.96 & 65.88 & 67.43 & 66.35 & 64.25 & 71.75 \\
en\_fantasycoref & 63.94 & 56.28 & 57.25 & 56.59 & 60.90 & 61.75 \\
en\_gum & 62.28 & 54.13 & 63.38 & 63.20 & 64.47 & 64.01 \\
en\_litbank & 61.44 & 56.86 & 63.77 & 61.64 & 62.97 & 64.28 \\
es\_ancora & 60.58 & 62.56 & 63.12 & 63.70 & 64.39 & 66.37 \\
fr\_ancor & 60.55 & 38.53 & 61.15 & 61.17 & 61.47 & 64.64 \\
fr\_democrat & 55.53 & 49.11 & 53.97 & 56.91 & 56.04 & 58.82 \\
fr\_litbankfr & 48.73 & 40.32 & 45.95 & 46.37 & 47.04 & 50.27 \\
grc\_proiel & 28.48 & 28.16 & 37.52 & 38.18 & 41.40 & 45.27 \\
hbo\_ptnk & 20.70 & 13.80 & 32.42 & 29.88 & 34.27 & 22.89 \\
hi\_hdtb & 60.58 & 54.51 & 53.22 & 54.00 & 52.49 & 64.57 \\
hu\_korkor & 34.37 & 37.43 & 36.41 & 35.52 & 37.63 & 39.62 \\
hu\_szegedkoref & 39.24 & 36.99 & 41.97 & 41.17 & 38.26 & 45.08 \\
ko\_ecmt & 57.38 & 49.64 & 55.75 & 55.50 & 56.02 & 57.41 \\
la\_coreflat & 4.93 & 9.72 & 7.44 & 5.29 & 0.90 & 0.00 \\
lt\_lcc & 50.33 & 51.49 & 48.19 & 49.16 & 48.62 & 55.10 \\
nl\_openboek & 56.02 & 43.14 & 55.65 & 57.56 & 56.87 & 56.72 \\
no\_bokmaalnarc & 61.73 & 50.93 & 60.01 & 60.97 & 62.25 & 66.35 \\
no\_nynorsknarc & 60.81 & 51.15 & 60.06 & 59.83 & 59.29 & 66.72 \\
pl\_pcc & 51.78 & 54.29 & 61.13 & 61.53 & 60.80 & 63.90 \\
ru\_rucor & 55.46 & 53.58 & 57.06 & 57.04 & 58.90 & 61.77 \\
tr\_itcc & 23.58 & 32.40 & 40.68 & 39.62 & 41.08 & 46.42 \\
\midrule
Average & 48.22 & 45.33 & 51.03 & 51.04 & 51.44 & 54.40 \\
\bottomrule
\end{tabular}
\caption{Coreference resolution system developmeent results. The model we used is \texttt{gemma-3-1b-it}, finetuned for 1 epoch on all datasets.}
\label{tab:development_set_formats_rsults}
\end{table*}

\begin{table*}[t!] 
\centering

\noindent\section{Mention Density and Prediction Error}
\label{sec:mention_density}

\setlength{\tabcolsep}{7pt}

\begin{tabular}{lccc}
\toprule
\multirow{2}{*}{Dataset} & \multicolumn{2}{c}{Mentions / 100 Tokens} & \multirow{2}{*}{\makecell{Relative Error\\(Pred vs Gold)}} \\
\cmidrule(lr){2-3}
 & Train Set (Gold) & Development Set (Predictions) & \\
\midrule
ca\_ancora & 14.63 & 12.61 & -0.14 \\
cs\_pcedt & 14.88 & 12.08 & -0.19 \\
cs\_pdt & 22.24 & 15.91 & -0.28 \\
cs\_pdtsc & 25.65 & 22.87 & -0.11 \\
cu\_proiel & 35.76 & 22.56 & -0.37 \\
de\_potsdamcc & 16.20 & 14.84 & -0.08 \\
en\_fantasycoref & 16.59 & 15.76 & -0.05 \\
en\_gum & 28.10 & 27.38 & -0.03 \\
en\_litbank & 13.91 & 14.59 & 0.05 \\
es\_ancora & 15.60 & 13.04 & -0.16 \\
fr\_ancor & 24.55 & 23.56 & -0.04 \\
fr\_democrat & 27.87 & 19.85 & -0.29 \\
fr\_litbankfr & 13.55 & 22.27 & 0.64 \\
grc\_proiel & 33.27 & 24.47 & -0.26 \\
hbo\_ptnk & 26.96 & 8.46 & -0.69 \\
hi\_hdtb & 18.10 & 13.43 & -0.26 \\
hu\_korkor & 16.85 & 10.22 & -0.39 \\
hu\_szegedkoref & 12.54 & 7.14 & -0.43 \\
ko\_ecmt & 25.23 & 23.09 & -0.09 \\
la\_coreflat & 7.65 & 0.15 & -0.98 \\
lt\_lcc & 12.38 & 7.56 & -0.39 \\
nl\_openboek & 23.39 & 19.96 & -0.15 \\
no\_bokmaalnarc & 30.27 & 29.54 & -0.02 \\
no\_nynorsknarc & 29.90 & 28.95 & -0.03 \\
pl\_pcc & 34.89 & 36.16 & 0.04 \\
ru\_rucor & 10.20 & 10.44 & 0.02 \\
tr\_itcc & 38.35 & 35.78 & -0.07 \\
\midrule
Mean & 21.83 & 18.25 & -0.16 \\
Min & 7.65 & 0.15 & -0.98 \\
Max & 38.35 & 36.16 & -0.06 \\
\bottomrule
\end{tabular}

\caption{Cross-dataset variation in mention density (per 100 tokens) and model prediction bias. The model we used is \texttt{gemma-3-1b-it}, finetuned for 1 epoch on all datasets.}
\label{}

\end{table*}

\begin{table*}[t!] 
\centering

\noindent\section{MiniDev set Dataset-Specific Adapter Results}
\label{sec:DatasetAdapter}

\setlength{\tabcolsep}{6pt}

\begin{tabular}{lccccccccc}
\toprule
\multirow{2}{*}{\vspace{-0.5em}Train Tok} & \multirow{2}{*}{\vspace{-0.5em}Dataset} & \multicolumn{6}{c}{Dataset-Specific Epoch} & \multirow{2}{*}{\vspace{-0.5em}{\makecell{Best\\Gain}}} & \multirow{2}{*}{\vspace{-0.5em}{\makecell{4 Epochs\\All Datasets}} }\\
\cmidrule(lr){3-8}
                            &                         & 0 & 1 & 2 & 3 & 4 & 5 & {} &\\
\midrule
7,727 & hbo\_ptnk &     22.89 & 12.69 & 57.21 & 56.15 & \textbf{57.84} & 56.94 &    34.95  & 53.26 \\
19,457 & hu\_korkor &   39.62 & 34.25 & 45.72 & 49.97 & \textbf{51.34} & 51.19 &    11.72 & 47.55 \\
20,726 & la\_coreflat & 0.00 & 0.00 & 13.15 & 28.11 & 29.38 & \textbf{34.02} &      34.02 & 18.32 \\
26,677 & de\_potsdam &  71.75 & 71.45 & 69.06 & \textbf{73.31} & 71.74 & 72.23 &    1.56 & 73.56 \\
30,082 & lt\_lcc &      55.10 & 57.29 & \textbf{71.67} & 71.20 & 69.86 & 69.43 &    16.57 & 60.15 \\
41,592 & hi\_hdtb & 64.57 & \textbf{73.41} & 71.48 & 72.42 & 71.71 & 70.66 &        8.84 & 72.78 \\
45,125 & tr\_itcc & 46.42 & 50.73 & \textbf{52.98} & 50.20 & 48.98 & 49.69 &        6.56 & 50.63 \\
47,853 & cu\_proiel & 31.62 & 40.26 & 43.87 & \textbf{45.50} & 44.81 & 44.60 &      13.88 & 41.55 \\
56,131 & grc\_proiel & 45.27 & 52.25 & 52.15 & 55.48 & 53.83 & \textbf{55.95} &     10.68 & 54.01 \\
57,322 & nl\_openboek & 56.72 & \textbf{58.41} & 57.94 & 58.01 & 57.18 & 57.46 &    1.69 & 58.98 \\
100,508 & hu\_szeged & 45.08 & 55.20 & 47.78 & 55.29 & \textbf{56.62} & 55.07 &     11.54 & 49.08 \\
123,599 & ru\_rucor & 61.77 & 62.67 & \textbf{65.02} & 64.94 & 64.15 & 62.96 &      3.25 & 64.05 \\
168,247 & en\_litbank & 64.28 & \textbf{69.16} & 67.93 & 68.13 & 67.79 & 67.96 &    4.88 & 69.66 \\
172,764 & no\_nynorsk & 66.72 & \textbf{69.99} & 68.46 & 65.28 & 68.05 & 66.85 &   3.27 & 68.92 \\
177,410 & en\_gum & 64.01 & 68.77 & \textbf{69.25} & 68.48 & 68.71 & 67.29 &        5.24 & 68.69 \\
195,869 & fr\_litbankfr & 50.27 & 50.71 & 51.64 & \textbf{53.40} & 51.74 & 50.22 &  3.13 & 54.89 \\
203,220 & no\_bokmaal & 66.35 & 69.64 & 67.18 & 67.73 & \textbf{69.73} & 66.59 &   3.38 & 69.45 \\
228,100 & fr\_democrat & 58.82 & \textbf{64.17} & 64.04 & 63.13 & 62.66 & 62.06 &   5.35 & 61.04 \\
275,491 & en\_fantasy & 61.75 & 67.80 & \textbf{68.09} & 67.11 & 67.86 & 67.78 &    6.34 & 68.07 \\
332,877 & ca\_ancora & 64.68 & 65.14 & 67.74 & \textbf{68.19} & 67.30 & 68.10 &     3.51 & 68.80 \\
366,903 & es\_ancora & 66.37 & 69.11 & 70.20 & \textbf{70.98} & 70.70 & 69.82 &     4.61 & 71.27 \\
371,775 & fr\_ancor & 64.64 & \textbf{70.83} & 68.75 & 68.08 & 66.94 & 68.51 &      6.19 & 67.56 \\
395,048 & ko\_ecmt & 57.41 & 58.89 & 59.45 & \textbf{59.76} & 58.64 & 57.80 &       2.35 & 58.01 \\
431,618 & pl\_pcc & 63.90 & 64.93 & \textbf{66.70} & 64.96 & 64.91 & 64.51 &        2.80 & 65.26 \\
614,217 & cs\_pdtsc & 60.26 & 60.02 & 61.95 & 61.88 & \textbf{63.18} & 62.62 &      2.92 & 62.62 \\
653,713 & cs\_pdt & 57.55 & \textbf{64.80} & 64.58 & 64.70 & 63.81 & 62.41 &        7.25 & 62.98 \\
935,568 & cs\_pcedt & 61.02 & 62.86 & 63.26 & \textbf{64.56} & 62.35 & 64.40 &      3.54 & 65.11 \\
\midrule
  & Average & 54.40 & 57.24 & 60.27 & \textbf{61.37} & 61.18 & 61.01 &              8.15 & 60.23 \\
\midrule
  & BestAdapter & 0 & 7 & 6 & 7 & 5 & 2 & - & -    \\
\bottomrule
\end{tabular}
\caption{Performance on the MiniDev set for dataset-specific LoRA adapters developed from the \texttt{gemma-3-1b-it} model that was finetuned for 1 epoch on all datasets. Column 0 corresponds to the multilingual adapter (no dataset-specific fine-tuning), while columns 1--5 report additional epochs of dataset-specific training. \textit{Best Gain} indicates the improvement over the multilingual baseline, and \textit{4 Epochs All Datasets} corresponds to uniform fine-tuning across all datasets for 4 epochs ; comparable to the 3 dataset-specific epochs configuration.}
\label{tab:dataset_specific_adapters}

\end{table*}

\begin{table}[t] 
\centering
\begin{minipage}{\columnwidth} 
\section{27B Parameters Training and Inference Configuration}
\label{sec:hyperparameters}

The following lists the hyperparameters used for the final 27B submission.

\paragraph{Base model}
\begin{itemize}[leftmargin=*] 
    \item Model: \texttt{google/gemma-3-27b-it}
    \item Quantization: 4-bit NF4 (QLoRA)
    \item Attention: FlashAttention-2
\end{itemize}

\paragraph{LoRA}
\begin{itemize}[leftmargin=*]
    \item Rank $r = 64$, $\alpha = 128$, dropout $= 0.01$
    \item Target modules: all linear layers
\end{itemize}

\paragraph{Training}
\begin{itemize}[leftmargin=*]
    \item Batch size: 1 (Accumulation: 32)
    \item Learning rate: 5e-5 (Linear)
    \item Sentences per batch: 6
    \item Context: 1,024 tokens
    \item Loss: completion-only CE
\end{itemize}

\paragraph{Inference}
\begin{itemize}[leftmargin=*]
    \item Sentences per batch: 6
    \item Context: 3,072 tokens
\end{itemize}
\end{minipage}
\end{table}

\clearpage

\begin{table*}[t] 
\centering

\noindent
\section{Detailed Results on the Test Set}
\label{sec:testset_final_results}

\renewcommand{\arraystretch}{1.1}
\resizebox{\textwidth}{!}{
\begin{tabular}{lccccccccccc}
\toprule
Track &\multicolumn{7}{c}{Unconstrained} &\multicolumn{4}{c}{LLM} \\
\cmidrule(lr){2-8} \cmidrule(lr){9-12}

\multirow{2}{*}{System} 
&\multicolumn{3}{c}{CorPipe} 
&\multirow{2}{*}{thmorton} 
&\multirow{2}{*}{Stanza} 
&\multirow{2}{*}{\makecell{Crac\\Baseline}} 
&\multirow{2}{*}{AU-KBC} 
&\multirow{2}{*}{\makecell{\textbf{Ours}\\\textbf{LatticeNLP}}}
&\multirow{2}{*}{hejmanj} 
&\multirow{2}{*}{portnlp} 
&\multirow{2}{*}{pavlk-mm} \\
\cmidrule(lr){2-4}

&Ensemble &Single &Single-lg & & & & & & & & \\
\midrule
ca\_ancora &\textbf{84.42} &83.04 &79.65 &76.77 &77.73 &67.91 &38.80 &77.02 &\ul{82.67} &73.68 &41.32 \\
cs\_pcedt &\textbf{79.34} &78.83 &73.85 &70.76 &74.00 &63.36 &25.77 &72.79 &\ul{75.83} &71.55 &36.66 \\
cs\_pdt &\textbf{81.81} &81.59 &77.93 &74.61 &76.40 &66.21 &36.77 &76.31 &\ul{80.03} &74.10 &40.31 \\
cs\_pdtsc &\textbf{76.70} &76.66 &72.30 &68.22 &71.22 &66.06 &37.91 &70.89 &\ul{73.85} &69.77 &40.93 \\
cu\_proiel &\textbf{68.75} &67.86 &56.27 &57.20 &38.92 &24.68 &34.80 &\ul{60.09} &59.58 &57.94 &29.35 \\
de\_potsdamcc &\textbf{75.00} &74.79 &69.92 &71.92 &70.36 &52.36 &29.77 &71.33 &\ul{73.35} &67.94 &49.26 \\
en\_fantasycoref &\textbf{81.12} &80.75 &74.92 &72.99 &69.90 &65.14 &41.35 &78.79 &\ul{80.20} &74.80 &58.07 \\
en\_gum &\textbf{77.42} &76.73 &73.48 &68.63 &72.69 &61.88 &44.23 &\ul{76.90} &75.94 &70.60 &55.30 \\
en\_litbank &\ul{85.33} &83.68 &79.38 &73.62 &73.32 &66.17 &34.41 &\textbf{85.40} &84.58 &78.37 &63.00 \\
es\_ancora &\textbf{85.28} &84.23 &82.30 &77.39 &80.14 &70.26 &37.35 &78.30 &\ul{83.07} &75.36 &47.27 \\
fr\_ancor &\ul{76.49} &75.45 &71.51 &70.18 &69.55 &61.82 &37.38 &77.33 &\textbf{77.51} &70.55 &46.42 \\
fr\_democrat &73.38 &\ul{73.80} &70.98 &68.99 &57.10 &55.60 &35.50 &\textbf{74.46} &66.66 &53.40 &20.72 \\
fr\_litbankfr &\textbf{82.46} &81.48 &76.52 &66.92 &64.52 &46.07 &28.55 &\ul{80.21} &60.47 &54.94 &46.66 \\
grc\_proiel &\textbf{79.01} &77.87 &69.05 &65.20 &53.86 &30.63 &43.36 &74.87 &\ul{75.28} &71.08 &43.43 \\
hbo\_ptnk &\ul{74.46} &72.00 &66.39 &57.90 &61.24 &31.70 &48.19 &\textbf{79.83} &76.80 &72.72 &61.40 \\
hi\_hdtb &\textbf{78.36} &77.83 &76.31 &72.82 &75.45 &66.60 &46.19 &76.84 &\ul{77.05} &75.61 &60.60 \\
hu\_korkor &\textbf{68.72} &68.23 &64.56 &60.52 &59.94 &42.24 &29.51 &\ul{65.47} &65.29 &59.15 &41.56 \\
hu\_szegedkoref &\textbf{72.61} &71.42 &67.71 &60.96 &66.87 &54.29 &31.65 &66.60 &\ul{69.00} &62.53 &37.33 \\
ko\_ecmt &\textbf{70.29} &69.72 &68.58 &61.07 &67.13 &64.97 &21.62 &\ul{68.84} &66.89 &69.48 &59.53 \\
la\_coreflat &\textbf{62.63} &58.69 &57.46 &55.95 &36.86 &6.80 &16.17 &\ul{58.66} &56.19 &44.79 &32.77 \\
lt\_lcc &\textbf{76.13} &75.18 &75.71 &66.49 &73.00 &62.42 &27.97 &65.35 &\ul{73.47} &68.08 &52.75 \\
nl\_openboek &\ul{74.72} &73.10 &69.88 &64.45 &60.03 &40.57 &34.51 &\textbf{77.42} &66.14 &72.93 &39.26 \\
no\_bokmaalnarc &\ul{78.90} &77.44 &74.42 &72.88 &72.84 &61.35 &42.93 &\textbf{81.19} &80.44 &72.13 &54.55 \\
no\_nynorsknarc &\ul{76.83} &76.44 &73.96 &72.02 &70.81 &61.09 &39.65 &77.59 &\textbf{79.45} &72.07 &53.49 \\
pl\_pcc &\textbf{82.07} &81.54 &78.04 &73.72 &73.68 &67.46 &36.25 &78.35 &\ul{80.06} &76.85 &43.17 \\
ru\_rucor &\textbf{86.21} &84.51 &82.15 &79.12 &80.38 &68.23 &33.72 &82.62 &\ul{84.65} &81.24 &52.29 \\
tr\_itcc &73.55 &\textbf{74.01} &69.52 &42.92 &60.93 &46.76 &37.21 &\ul{73.09} &69.09 &62.97 &39.77 \\
\midrule
Average &\textbf{77.11} &76.18 &72.32 &67.56 &67.00 &54.54 &35.24 &\ul{74.32} &73.83 &68.69 &46.19 \\
Standard Dev. &5.68 &5.89 &6.41 &8.04 &10.85 &15.98 &7.34 &6.59 &7.85 &8.44 &10.44 \\
\bottomrule
\end{tabular}
}
\caption{Official coreference resolution performance (CoNLL-F1) across all test sets. Overall best results in bold, track-best underlined (if not already bold).}
\label{tab:testset_final_results}
\end{table*}

\end{document}